\documentclass[letterpaper, 10 pt, conference]{ieeeconf}  

\IEEEoverridecommandlockouts                              

\usepackage{algorithmic}
\usepackage{amsmath} 
\usepackage{amssymb}  
\usepackage{graphicx}
\usepackage{bm}
\usepackage[ruled,vlined]{algorithm2e}
\usepackage{multirow}
\newcommand{\etal}{\textit{et al}.}

\DeclareMathOperator*{\argmax}{arg\,max}

\newcommand{\vu}{{\bf u}}
\newcommand{\vb}{{\bf b}}
\newcommand{\vv}{{\bf v}}

\title{\LARGE \bf
Object-Augmented RGB-D SLAM for Wide-Disparity Relocalisation
}

\author{Yuhang Ming, Xingrui Yang and Andrew Calway
\thanks{The authors are with the Visual Information Laboratory, Department of Computer Science,
        University of Bristol, Bristol, U.K.
        {\tt\small \{yuhang.ming, x.yang, andrew.calway\}@bristol.ac.uk}}%
\thanks{The source code is available at https://github.com/YuhangMing/Object-Guided-Relocalisation}
}

\begin{document}

\maketitle
\thispagestyle{empty}
\pagestyle{empty}

\begin{abstract}
We propose a novel object-augmented RGB-D SLAM system that is capable of constructing a consistent object map and performing relocalisation based on centroids of objects in the map. The approach aims to overcome the view dependence of appearance-based relocalisation methods using point features or images. 
During the map construction, 
we use a pre-trained neural network to detect objects and estimate 6D poses from RGB-D data. 
An incremental probabilistic model is used to aggregate estimates over time to create the object map. 
Then in relocalisation, we use the same network to extract objects-of-interest in the `lost' frames. 
Pairwise geometric matching finds correspondences between map and frame objects, and probabilistic absolute orientation followed by application of iterative closest point to dense depth maps and object centroids gives relocalisation. 
Results of experiments in desktop environments demonstrate very high success rates even for frames with widely different viewpoints from those used to construct the map, significantly outperforming two appearance-based methods.
\end{abstract}

\section{Introduction}
\label{sec:intro}

Vision based simultaneous localisation and mapping (SLAM) using RGB, RGB-D or stereo cameras is a key capability for autonomous robots and vehicles, and applications such as augmented reality (AR). Significant advances have been made, resulting in systems capable of real-time 3D tracking and scene reconstruction, operating over wide areas, both indoors and outdoors. However, more challenging has been the development of fast and reliable relocalisation, i.e., determining the 6D pose of a `lost' sensor, caused either by tracking failure or returning to a previously constructed map, a capability that is critical for robust operation.

The majority of approaches to relocalisation are based on appearance, seeking to align a lost frame with a map by matching sets of 2D or 3D  point features \cite{Williams2007, Chekhlov2008, Mur2014BoW, Li2015Pairwise} or by learning the relationship between pose and frame appearance \cite{Shotton2013SCoRe, Glocker2013FERNS, Gee2012}. These can be effective in feature-rich environments, but often struggle in featureless areas or when the lost frame viewpoint is significantly different from those used to generate the map. Although recent work using deep neural networks has increased robustness to appearance changes by learning to match 3D point clouds 
\cite{Yew20183DFeatNet, Lu2019DeepVCP, Du2020DH3D, Huang2021Predator}
, for example, they still rely on having sufficient intersection of feature sets.
Wide disparities cause problems and in extreme cases in which map features are not visible or frame appearance is significantly different, relocalisation will fail. This is the issue that we seek to address.

We adopt a semantic object based approach. In contrast to view-dependent point features, the presence of visible objects is invariant to viewpoint and recent advances in object detection \cite{He2017maskrcnn, Redmon2018Yolov3} and object pose estimation \cite{Wang2019NOCS, Qi2019deep} using deep neural networks gives potential for object based relocalisation capable of operating over wide disparities.
The main challenges for doing so then lie in 1) 
modelling the noise in the network detections to build a stable object map and 2) achieving data association between objects in the map and 
those in the lost frames. We address both of these issues in this work.

\begin{figure}[t]
\centering
\includegraphics[width=0.47\textwidth]{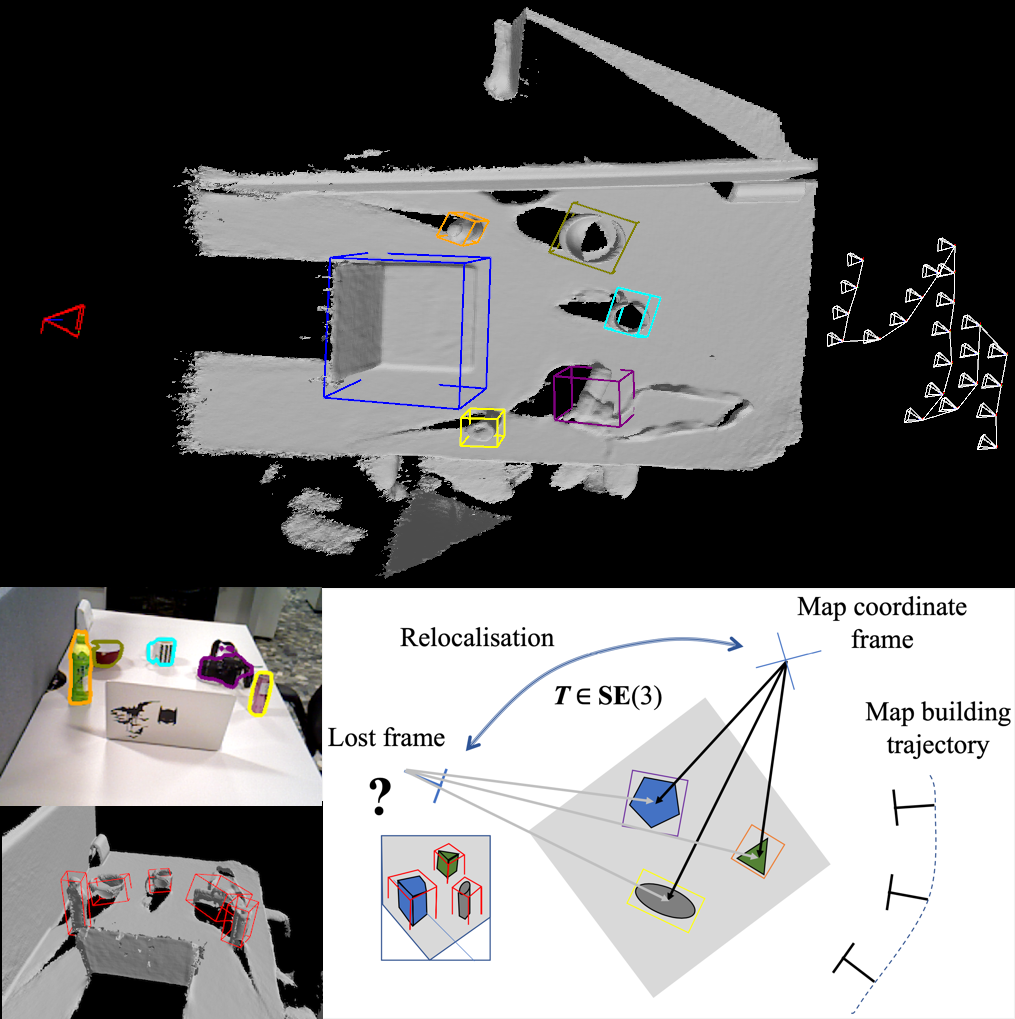}
\caption{Object based relocalisation: 
(top) a global dense map augmented by objects in bounding boxes along with a map construction trajectory and a lost frame; 
(left-middle) objects-of-interest detected in the lost frame; 
(left-bottom) objects bounding boxes estimated in the lost frame;
(right-bottom) the transformation $\bm{T}\in\bm{SE}(3)$ between the map and the lost frame yielded by aligning objects centroids using probabilistic absolute orientation.
}
\label{fig:idea}
\vspace*{-3ex}
\end{figure}

\begin{figure*}[t]
  \centering
  \includegraphics[width=0.98\textwidth]{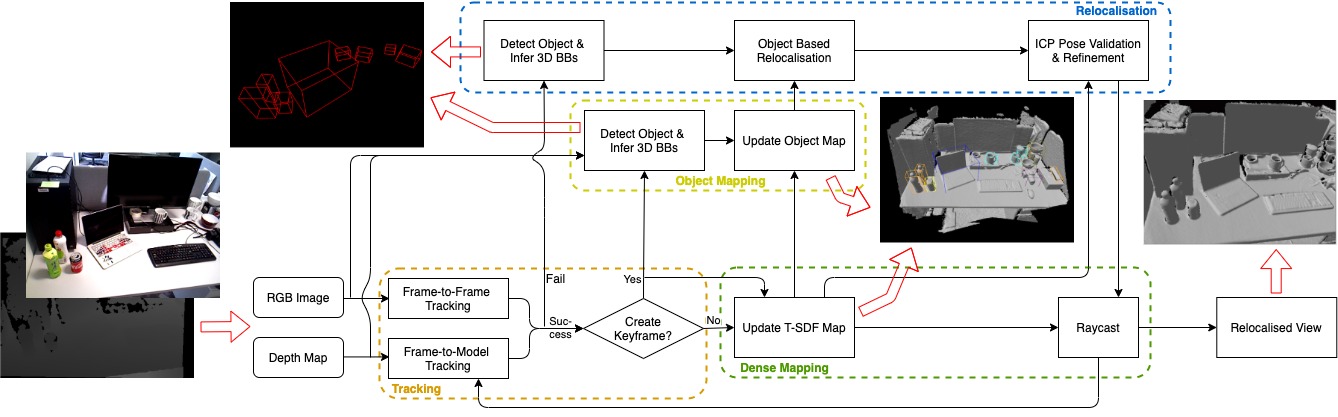}
  \caption{Overview of the main components and data flow within our system}
  \label{fig:overview}
  \vspace*{-3ex}
\end{figure*}

Our object-augmented RGB-D SLAM system is illustrated in Fig. \ref{fig:idea}. The system integrates an object map within a standard dense mapping framework, with objects represented by bounding boxes (BB) and associated centroids. The latter are detected in key frames using the single frame Normalised Object Coordinate Space (NOCS) model described in \cite{Wang2019NOCS} and re-detections in subsequent key frames are integrated over time. Crucially, to account for errors in the detections returned by the NOCS model, we allow each object to be represented by a finite set of overlapping BBs, with each centroid assumed to be normally distributed. The result is a stable object map sitting within the dense map. When relocalisation is required, we detect and estimate object poses in the lost frame and solve the data association by exploiting the pairwise geometry between objects. Finally, we use a coarse-to-fine optimisation to perform the relocalisation based on both object and geometric information.

Experiments were performed on 10 desktop scenes with 
varying degrees of content arrangement and clutter, for lost frames taken from widely disparate viewpoints, both close to and far from that used to construct the dense scene map. 
We compared performance with two other approaches, based on bag-of-words \cite{Mur2014BoW} and randomised ferns \cite{Glocker2013FERNS}, and demonstrate that the proposed method significantly outperforms both methods, especially for viewpoints with wide disparity relative to that used to generate the dense map.


\section{Related Work}
\label{sec:bg}
Considerable work has been done on using semantic and object information in visual SLAM \cite{Salas2013slam++, Sunderhauf2017semanticmapping, Runz2018maskfusion, Bowman2017dataassociation, Mccormac2018fusion++, Nicholson2019Quadricslam, Sucar2020Nodeslam}, as well as using deep learning to improve feature matching for robust relocalisation 
\cite{Yew20183DFeatNet, Lu2019DeepVCP, Du2020DH3D, Huang2021Predator, Sarlin2019coarsetofine, Stumberg2020GNNet, Sarlin2021back2features}
. However, less work has been done on using objects for relocalisation. The few examples include Fusion++ \cite{Mccormac2018fusion++}, which uses object detections to direct 3D point matching, and  SLAM++  \cite{Salas2013slam++}, which uses spatial configurations of known objects. The latter has similarities with our method, but differs in that it requires full 3D object models, created offline, to allow robust matching to depth data. 

In a different context, 3D spatial configuration of objects have been used for place recognition. For example, in \cite{Frampton2013placerecog}, spatial arrangement of objects such as road signs and bollards are used to match views to a database of place reference views, providing a degree of invariance to viewing direction and appearance. And in \cite{Finman2014physicalwords}, constellations of objects represented by physical words are matched to achieve place recognition in RGB-D SLAM. In both these works, matching of object configurations provides a coarse estimate of the relative pose of the two views obtained via absolute orientation (AO), in a similar fashion to our approach.

More recently, Yang \etal\ solve relocalisation w.r.t dense maps by incorporating semantic labels into the key frame encoding through randomised ferns \cite{Yang2019semanticreloc}, whilst for sparse SLAM, Li \etal\ describe a series of object based relocalisation and loop closure algorithms \cite{Li2018scenemodel, Li2019semanticmapping, Li2020loopclosure}. The latter are the closest previous works to our own and are also based on using bounding boxes (BBs) to represent object pose. However, unlike our method that is based on the centroids, they use the top-centre point of BBs to derive the pose of lost frames. Another key difference from our work is the method used to derive object bounding boxes, which are initialised and then sequentially updated based on intersections of their projections with 2D object detections. This means that matching only occurs once sufficient numbers of frames have been processed to build up the object representations, resulting in delayed relocalisation. This is in contrast to the single frame relocalisation that our method achieves.

\section{System Overview}
\label{sec:overview}

Our system consists of four main components - tracking, dense mapping, object mapping and relocalisation. The data flow and interactions are shown in Fig. \ref{fig:overview}. Our core dense SLAM engine is a modification of InfiniTAM \cite{Prisacariu2017infinitamv3}, with tracking implemented using iterative closest point (ICP) based on both geometric and photometric error. We use frame-to-model minimisation for the former as in \cite{Prisacariu2017infinitamv3} but use frame-to-frame minimisation for the latter to mitigate against distortions in the rendered texture map as in \cite{Whelan2015elasticfusion}. Key frames are generated according to camera movement and we construct a dense map by fusing RGB-D frames within a truncated signed distance function (TSDF) \cite{curless1996sdf} representation, using voxel hashing for efficient implementation \cite{niebner2013voxelhashing}.

The two key components are the object mapping and relocalisation. For the former, we make use of the NOCS network recently introduced by Wang \etal\ \cite{Wang2019NOCS}, which provides estimates of 6D pose and scale for object categories from a single RGB-D frame in addition to 2-D labels, masks and confidence scores obtained from Mask R-CNN \cite{He2017maskrcnn}. This means that for each RGB-D frame we have a set of 2-D object detections and for each object a 3D BB.
We use the pre-trained NOCS network, which is capable of detecting six different categories of objects (\textit{bottle, bowl, camera, can, laptop,} and \textit{mug}), and apply it to every key frame, with object detections instantiating new objects or updating existing objects in the map, to 
yield a consistent object representation.
Details are provided in Section \ref{sec:object-mapping}.

When an RGB-D frame is lost, as indicated by ICP failure, caused by erratic motion or if the sensor has entered a previously constructed map, for example, then relocalisation is started. 
We first perform object detection and pose estimation using the same NOCS network, as in the object mapping process. Then we match and align the centroids of objects detected in the lost frame with  
those already in the map to provide an estimate of the frame pose. 
The former is done by exploiting the pairwise geometry between objects and the later follows coarse-to-fine scheme basing on a combination of
probabilistic AO estimation and application of depth-centroids ICP to refine the pose estimate. Details are provided in Section \ref{sec:reloc}.


\section{Object Mapping}
\label{sec:object-mapping}
Relocalisation relies on building an object map w.r.t the system coordinate frame and reliably detecting objects and estimating their 3D positions and extent in `lost' frames, so they can be matched with the object map. As noted above, we do this by representing objects by 3D BBs and their associated centroids, estimated by the NOCS network. An alternative to using NOCS would be to use 2D BBs detected in key frames and then use depth measurements within those BBs integrated over time to determine 3D centroids and BBs. However, we found this to give inconsistent estimates, especially for large asymmetric objects, because single lost frames only provides surface depth information, thus precluding single frame relocalisation. 



\subsection{Single Frame BB Estimation}
\label{subsec:singleBB}

Given an RGB-D frame, the NOCS network provides detections of objects $O^f=\{o^f_0,o^f_1,\ldots,o^f_M\}$ from known categories, along with a category label $l^f_i$ and a coordinate map for each object. The latter is then aligned with the object depth map obtained from the RGB-D frame to give estimates of its centroid $\vu^f_i$ and associated BB $\vb^f_i$ defined by a 3D orientation $R^f_i$ and scale factor $s^f_i$. Alignment is achieved using the AO algorithm \cite{Horn1988AO} combined with random sample consensus (RANSAC). Full details can be found in \cite{Wang2019NOCS}.

\begin{figure}[t]
\centerline{
\begin{tabular}{cc}
\includegraphics[width=0.22\textwidth]{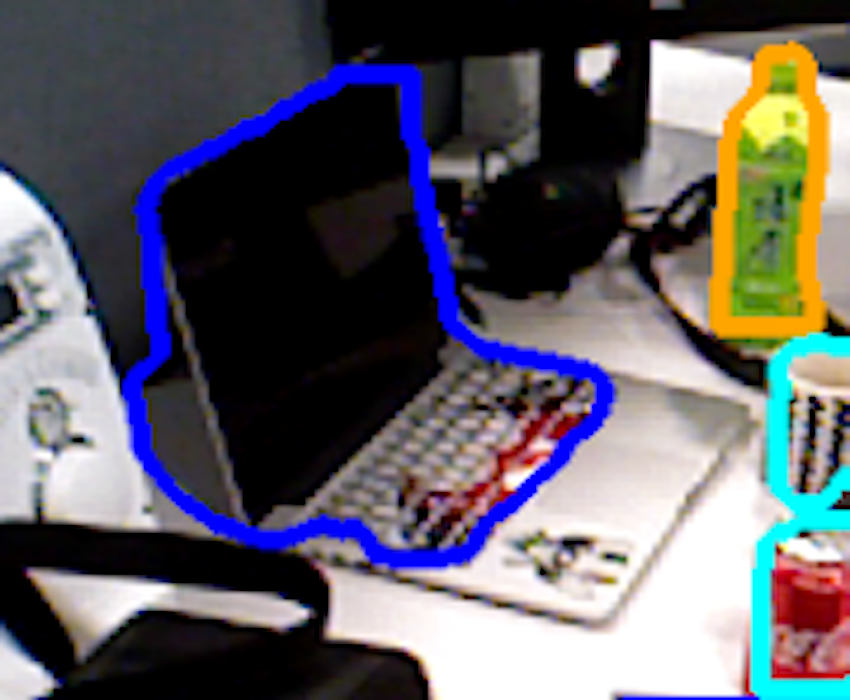}&
\includegraphics[width=0.22\textwidth]{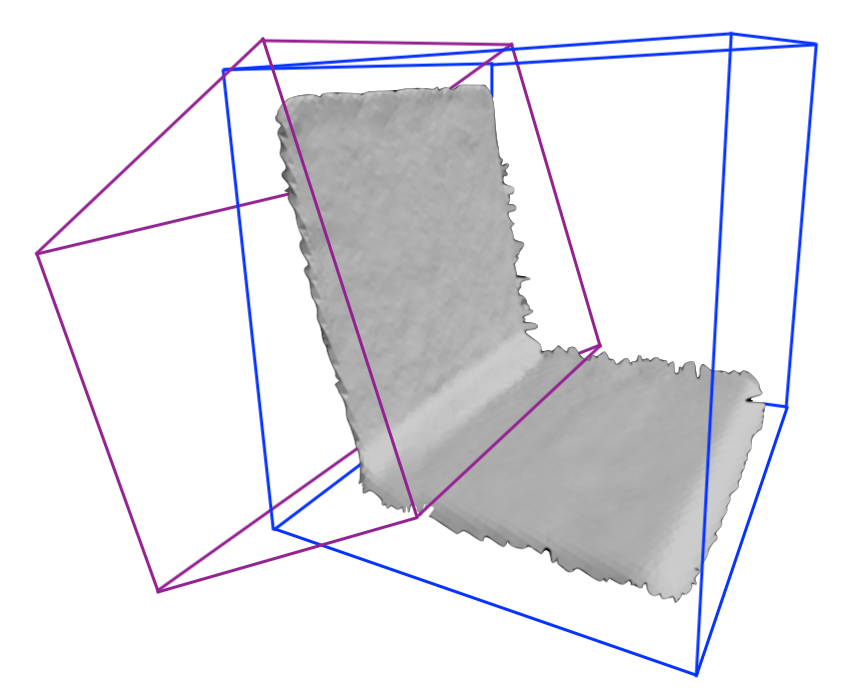}
\\
(a) & (b)\\
\includegraphics[width=0.22\textwidth]{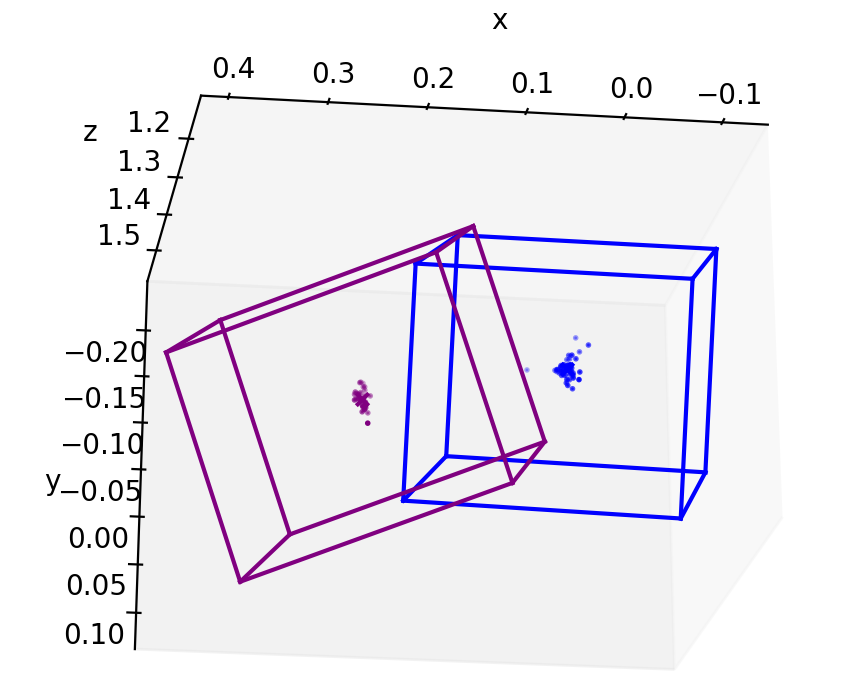}&
\includegraphics[width=0.22\textwidth]{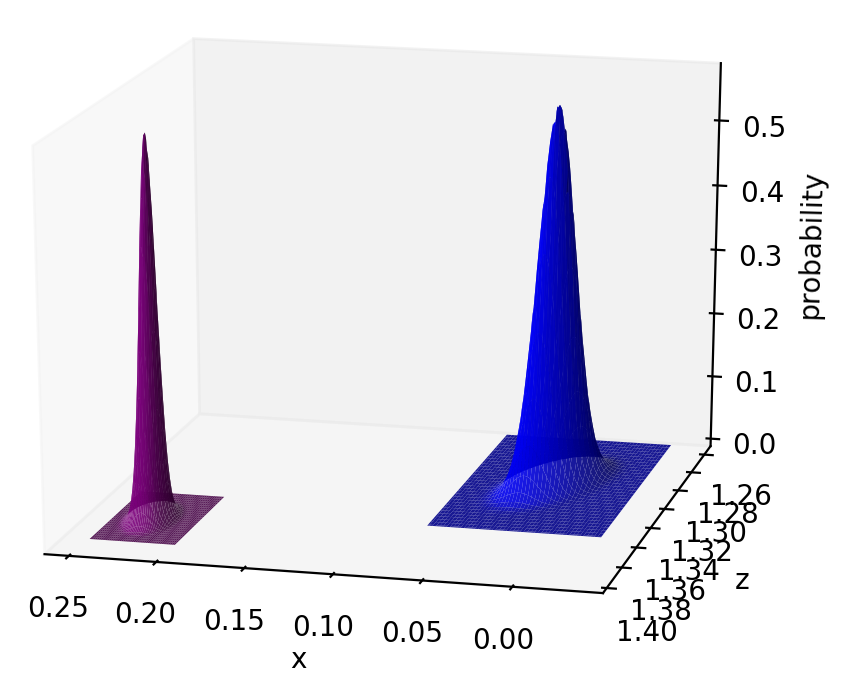}
\\
(c)&(d)\\
\end{tabular}}
\caption{
An example of a laptop instance, represented by two BBs with corresponding centroids, assumed to be normally distributed.
(a) laptop detection in RGB frame;
(b) flipped BBs aligned with the lid and keyboard (that in blue resulted from the detection in (a));
(c) centroid estimates returned from all assigned detections;
(d) normal distributions representing centroids, visualised by projecting into the x-z plane.
}
\label{fig:obj-rep}
\vspace*{-3ex}
\end{figure}

It is important to account for errors in the NOCS detections and estimates. In addition to missing detections (false negatives) and incorrect labelling of detected objects (false positives), we noted that the network sometimes returns inconsistent BB poses or `flipped configurations', typically caused by the alignment with different dominant features (see Fig. \ref{fig:obj-rep}b). These result from inconsistent predictions of the coordinate map, but also from errors in the object depth maps, such as holes caused by reflections, for example.

To address this, we represent each object with one or more BB configurations, with the respective centroid positions assumed to be normally distributed to account for the small errors observed for each configuration. To deal with false positives and negatives, we require an object to be detected in multiple successive key frames before being inserted into the map. Details of the representation and how objects are sequentially added to the map are given below.

\subsection{Sequential BB Fusion}
\label{subsec:BBfusing}

We assume the scene contains multiple objects, an unknown number $N$ of which are from $K$ known categories with labels $L=\{1,2,\ldots,K\}$, i.e., $K=6$ when using the pre-trained NOCS network. The object map is denoted by $O^m=\{o^m_0,o^m_1,\ldots,o^m_N\}$. 
Associated with each object $o^m_j$ is a category label $l^m_j\in L$ and one or more 
BB configurations  $(\bar{\vb}^m_{j0}\ldots\bar{\vb}^m_{jN_j})$ with associated centroids which are assumed to be normally distributed, i.e. $(\mathcal{N}(\bar{\vu}^f_{i0}, \Sigma^f_{i0})\dots\mathcal{N}(\bar{\vu}^f_{iN_j}, \Sigma^f_{iN_j}))$. An example of the representation is shown in Fig. \ref{fig:obj-rep}.

We use a loosely coupled integration between the densely reconstructed TSDF map and the sparse object map, sequentially updating and refining the object BB configurations based on the tracking and key frames from SLAM. Specifically, objects are initialised using estimates from the first key frame, 
i.e., given $M$ detected objects, $O^m=O^f$.
As subsequent key frames are created, we use the 6D pose produced in the tracking module to place detected frame objects within the system coordinate frame. We then associate corresponding objects from the key frame and the map by computing the intersection over union (IoU) metric between their respective BBs. Specifically, object $o^f_i$ detected in a key frame is assigned to and used to update map object $o^m_{j^*}$ if 
\begin{equation}
o^f_i\Rightarrow o^m_{j^*} \;\;\;
\mbox{iff}\;\;\; U(\vb^f_i,\bar{\vb}^m_{j^*k^*}) > \tau
\end{equation}
such that $(j^*,k^*)$ are defined as
\begin{equation}
(j^*,k^*)= \argmax_{\{(j,k) \;|\; l^f_i=l^m_j\}} U(\vb^f_i,\bar{\vb}^m_{jk})
\label{eqn:object_match_test}
\end{equation}
where $\tau$ is a threshold and $U(\vb_1,\vb_2)$ denotes the IoU metric for the BBs defined by $\vb_1$ and $\vb_2$, i.e., we assign estimated objects to map objects if they have the same category label and have sufficient overlap between their BBs. 


Given a pair of matched objects $o^f_i$ and $o^m_{j^*}$, we determine which BB configuration should be updated by testing the squared Mahalanobis distance (MD) $d_k^2=(\vu^f_i-\bar{\vu}^m_{j^*k})^T({\Sigma^m_{j*k}})^{-1}(\vu^f_i-\bar{\vu}^m_{j^*k})$ between the new centroid estimate $\vu^f_i$ and the mean centroids of each configuration $\bar{\vu}^m_{j^*k}$ 
against the chi-squared distribution $\chi^2_{\nu,\alpha}$ (we used the degree of freedom $\nu=3$ and the critical value $\alpha = 0.001$).

If $d_k^2>\chi^2_{\nu,\alpha}$ for all $k$, then we initialise a new BB for the object. If $d_k^2<\chi^2_{\nu,\alpha}$ for only one configuration, say $k=k'$, then we update $\mathcal{N}(\bar{\vu}^m_{j^{*}k'}, \Sigma^m_{j^{*}k'})$, i.e., replacing $\bar{\vu}^m_{j^{*}k'}$ and $\Sigma^m_{j^*k'}$ with the average and covariance of previously assigned and new centroids, and $\bar{\vb}^m_{j^*k'}$ with the average of previously assigned and new BBs, which is done over 3D orientations and scales. If the squared MD test is satisfied for two or more configurations, then we merge these into a single configuration and initialise using the previously assigned and new centroid estimates. 

If no matching map object is found, then a new map object is initialised as described above. To deal with false positive and negative detections, we require that in order to persist in the map, initialised objects must obtain updates from a sufficient number of successive key frames within which they are expected to be in view. In the experiments we used short finite length sequences to generate the object maps and required that objects receive updates from at least 25\% of relevant key frames. Note that categorisation errors could result in multiple separate object representations for the same physical object. Although we observed very few such cases in the experiments, this turns out to be a desirable feature for relocalisation, since it will likely account for any miss-classification errors in lost frames.


\section{Relocalisation}
\label{sec:reloc}

Given a lost frame, we want to determine its pose within the system coordinate frame by aligning object estimates from the NOCS network with those in the object map. We, therefore, have to first solve the correspondence problem, matching frame objects to map objects, and then determining the transformation $\bm{T}\in\bm{SE}(3)$ which best aligns the two sets. We base the latter on aligning centroids as opposed to complete 6D poses, as we found that the estimates of 3D orientation from the NOCS network are not sufficiently consistent. This means that we require at least 3 matching objects to determine ${\bm T}$. 

Our relocalisation process consists of three stages. First, we use a graph matching algorithm to determine a set of correspondences based on pairwise geometry \cite{Leordeanu2005EigenPair}\cite{Li2015Pairwise}, followed by application of probabilistic AO coupled with RANSAC to determine an initial estimate of frame pose. Finally, this is then refined and validated by applying a novel object-incorporated ICP to align the lost frame depth values and the object centroids with those in the map. 


\subsection{Correspondence Matching using Pairwise Geometry}
\label{subsec:pairwise}

Consider that we have $M$ object estimates $O^f$ in a lost frame and $N$ objects $O^m$ in the map. Each frame object has an associated centroid $\vu^f_i$ and each map object has $N_j$ configuration centroids $\vu^m_{jk}$. Motivated by \cite{Leordeanu2005EigenPair} and \cite{Li2015Pairwise}, we seek a set of correspondences between $O^f$ and $O^m$ such that each correspondence is sufficiently similar and that pairs of correspondences are consistent in terms of their geometric relationship, subject to a one-to-one mapping constraint.

Finding the optimal set of such correspondences is NP-hard and so we adopt an approximation using the spectral technique described in \cite{Leordeanu2005EigenPair}. This is based on the principal eigenvector of the square adjacency matrix $A$ representing the relationship between pairs of potential correspondences, the number of which defines the the size of $A$. The component $A_{ii}$ represents the similarity within the $i$th correspondence and $A_{ij}$ represents the geometric consistency between the $i$th and $j$th correspondences. In our case, we first identify all potential correspondences based on category labels, i.e., all centroid pairs of the form $(\vu^f_i, \vu^m_{i'k_{i'}})$ such that $l^f_i=l^m_{i'}$, and then define $A$ as
\begin{equation}
    A_{ij} = 
    \begin{cases}
        \mbox{min}(s^f_i/s^m_{i'k_{i'}}, s^m_{i'k_{i'}}/s^f_i) & \text{if } i=j\\
        \exp{-|d^f_{ij}-d^m_{i'j'}|} & \text{otherwise}
    \end{cases}
\label{eqn:adjmat}
\end{equation}
where $d^f_{ij}=||\vu^f_{i}-\vu^f_{j}||_2$ and $d^m_{i'j'}=||\bar{\vu}^m_{i'k_{i'}}-\bar{\vu}^m_{j'k_{j'}}||_2$ are the Euclidean distances between the centroids in the frame and map, respectively, and $s^f_i$ and $s^m_{i'k_{i'}}$ are the scales of the BBs associated with the frame and map centroids, respectively. In other words, we favour matches with similar size BBs and seek consistency between the relative 3D location of objects estimated in the frame and that of corresponding object configurations in the map.

The components of the principal eigenvector of $A$ provide a ranking of correspondences in terms of the strength of their association with the main cluster of nodes in the graph defined by $A$, i.e., correspondences with high individual and pairwise similarity. As in \cite{Leordeanu2005EigenPair}, we use this to find the subset of correspondences which maximises the sum of association scores whilst maintaining one-to-one mapping. This then forms the set of selected correspondences $\mathcal{D}$.

\subsection{Probabilistic Absolute Orientation}
\label{subsec:probAO}
Given the list of best correspondences $\mathcal{D}$, we now compute an initial estimate of the 6D pose of the lost frame using probabilistic AO \cite{Segal2009GenICP}. However, as reported in \cite{Leordeanu2005EigenPair}, outliers may still exist in the $\mathcal{D}$. Therefore, RANSAC is applied to rule out the outliers, using the minimal set of 3 correspondences in each step. Given a subset of correspondences from $\mathcal{D}$, denoted $\mathcal{D}'$, each defined by a frame centroid $\vu^f_i$ and a corresponding mean centroid $\bar{\vu}^m_{i'k_{i'}}$ in the map, a pose estimate for the lost frame is then given by
\begin{equation}
    \mathbf{T}^* = \underset{\mathbf{T}}{\text{ argmin }} \sum_{\mathcal{D}'} 
    \bm{d}_i(\mathbf{T})^T
    (\Sigma^m_{i'k_{i'}})^{-1}
    \bm{d}_i(\mathbf{T})
    \label{eqn:prob_AO}
    \vspace*{-1ex}
\end{equation}
where $\bm{d}_i(\mathbf{T}) = \bar{\vu}^m_{i'k_{i'}} - \vv(\vu^f_i,\mathbf{T})$ is the distance between the corresponding centroids following transformation and $\vv(\vu,\mathbf{T})$ denotes the transformation of $\vu$ by $\mathbf{T}$. This can be solved by iterative optimisation starting from an initial estimate obtained using a closed form solution for standard AO, i.e., corresponding to setting $\Sigma^m_{i'k_{i'}}=I$ in (\ref{eqn:prob_AO}). Following completion of RANSAC iterations, final estimation over all inliers provides an initial estimate of the frame pose $\mathbf{T}^*_{init}$.

\subsection{Depth-Centroid ICP}
\label{subsec:icp}
Once the initial pose estimate is obtained from the probabilistic AO, the pose is then validated and refined using ICP. We base the latter on a loss function incorporating both the corresponding centroids found in the previous step and the set of 3D points obtained from the RGB-D frame and reconstructed dense map. Relying purely on the latter risks converging on a poor pose estimate due to errors in the frame depth estimates and the dense map, especially within the latter if it is incomplete, which is a scenario that we wish to address by using object based matching. Let $\Lambda$ denote the set of 3D points ${\bf p}^f_j$ corresponding to valid depth estimates in the RGB-D frame and $\mathcal{D}_{in}$ the set of inlier correspondences in $\mathcal{D}$, then the final estimate of the lost frame pose $\mathbf{T}^*$ is found by iterative optimisation of the following point-to-plane ICP problem, starting from the initial estimate $\mathbf{T}^*_{init}$
\begin{equation}
\begin{split}
    \mathbf{T}^* = \underset{\mathbf{T}}{\text{ argmin }} \, &\omega_1 \sum_\Lambda ((\bm{n}^m_{j'})^T(\mathbf{p}^m_{j'} - \vv(\mathbf{p}^f_j,\mathbf{T}))^2\\
    &+ \;\; \omega_2 \sum_{\mathcal{D}_{in}} || \bar{\vu}^m_{i'k_{i'}} - \vv(\vu^f_i,\mathbf{T}) ||^2
\end{split}
\end{equation}
where $\mathbf{p}^m_{j'}$ is the closest map point to $\vv(\mathbf{p}^f_j,\mathbf{T})$ and $\bm{n}^m_{j'}$ is the surface normal at $\mathbf{p}^m_{j'}$. Weights $\omega_1$ and $\omega_2$ control the contribution of each term. Intuitively speaking, the larger the disparities are, the more contribution should the object term make. But we found that setting $\omega_1$ = $\omega_2=1$ gave good results across different scenes.

\section{Experiments}
\label{sec:evaluation}

\begin{figure*}[t]
\centerline{
\begin{tabular}{|ccccc|}
\hline
Scene05 
& (Horizontal view change)
& \multicolumn{3}{l|}{Average relocalisation error: 2.30cm, $1.51^{\circ}$.}
\\
\multicolumn{2}{|c}{\multirow{6}{17em}{ \includegraphics[width=0.35\textwidth]{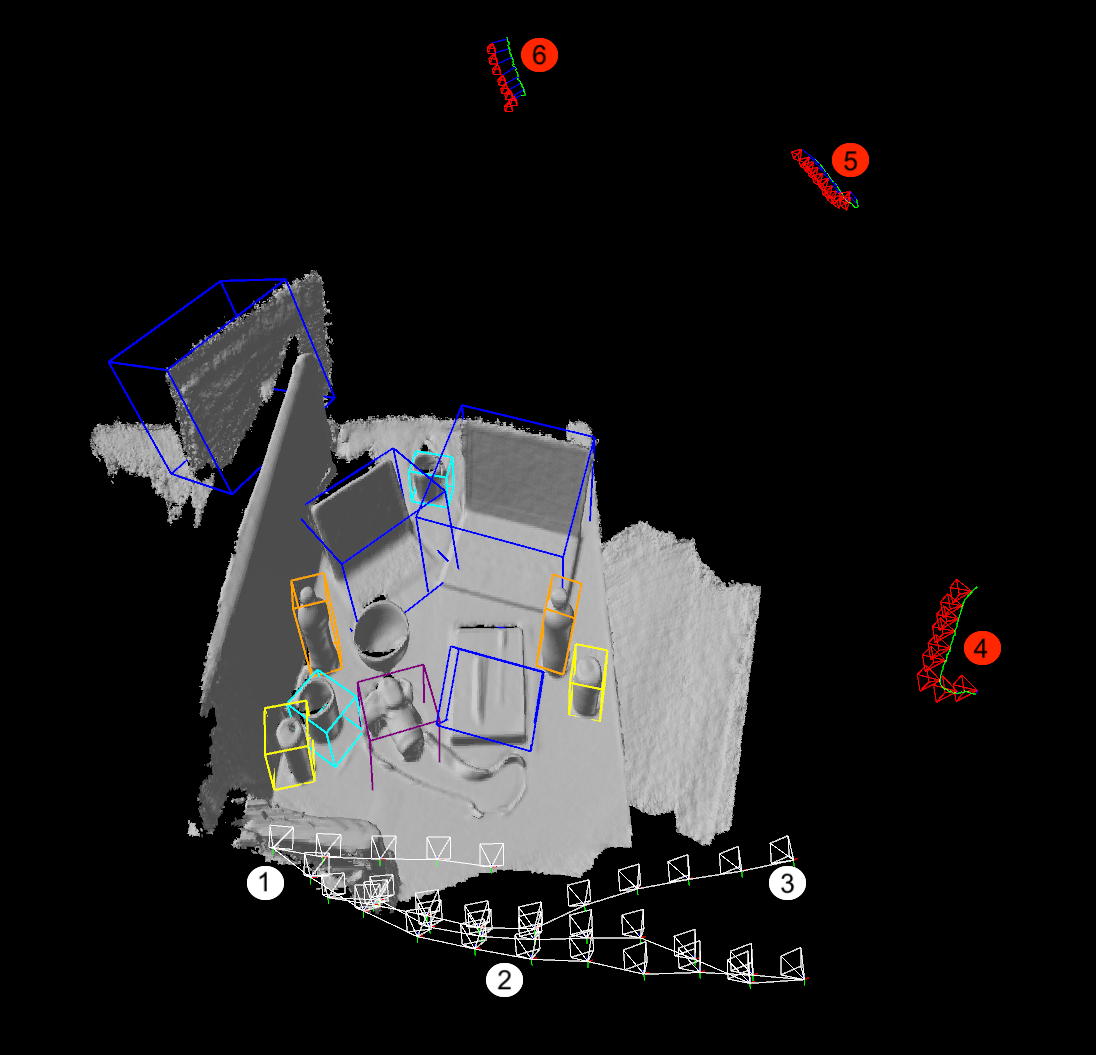} }}
& \multicolumn{3}{l|}{Map construction segment (MCS)} 
\\
\multicolumn{2}{|c}{}
&\includegraphics[width=0.18\textwidth]{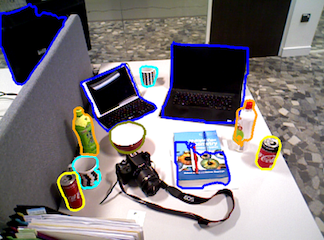}
&\includegraphics[width=0.18\textwidth]{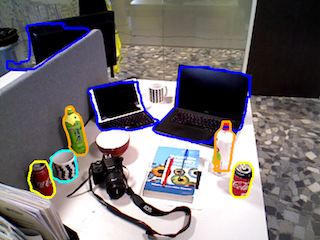}
&\includegraphics[width=0.18\textwidth]{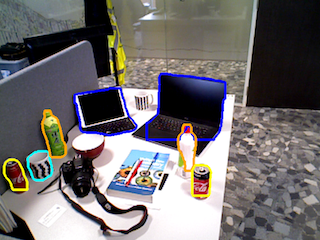}
\\
\multicolumn{2}{|c}{}
&Example 1
&Example 2
&Example 3
\\
\multicolumn{2}{|c}{}
& \multicolumn{3}{l|}{Relocalisation segment (RS)} 
\\
\multicolumn{2}{|c}{}
&\includegraphics[width=0.18\textwidth]{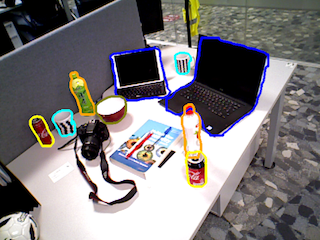}
&\includegraphics[width=0.18\textwidth]{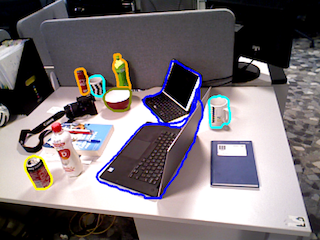}
&\includegraphics[width=0.18\textwidth]{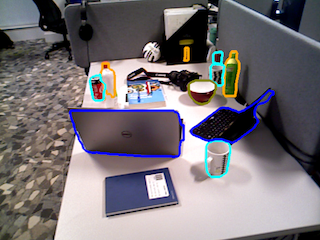}
\\
\multicolumn{2}{|c}{}
&Example 4 ($30^{\circ}(h)$)
&Example 5 ($120^{\circ}(h)$)
&Example 6 ($180^{\circ}(h)$)
\\
\hline
\hline
Scene07 
& (Vertical view change) 
& \multicolumn{3}{l|}{Average relocalisation error: 1.56cm, $1.58^{\circ}$.}
\\
\multicolumn{2}{|c}{\multirow{6}{17em}{ \includegraphics[width=0.35\textwidth]{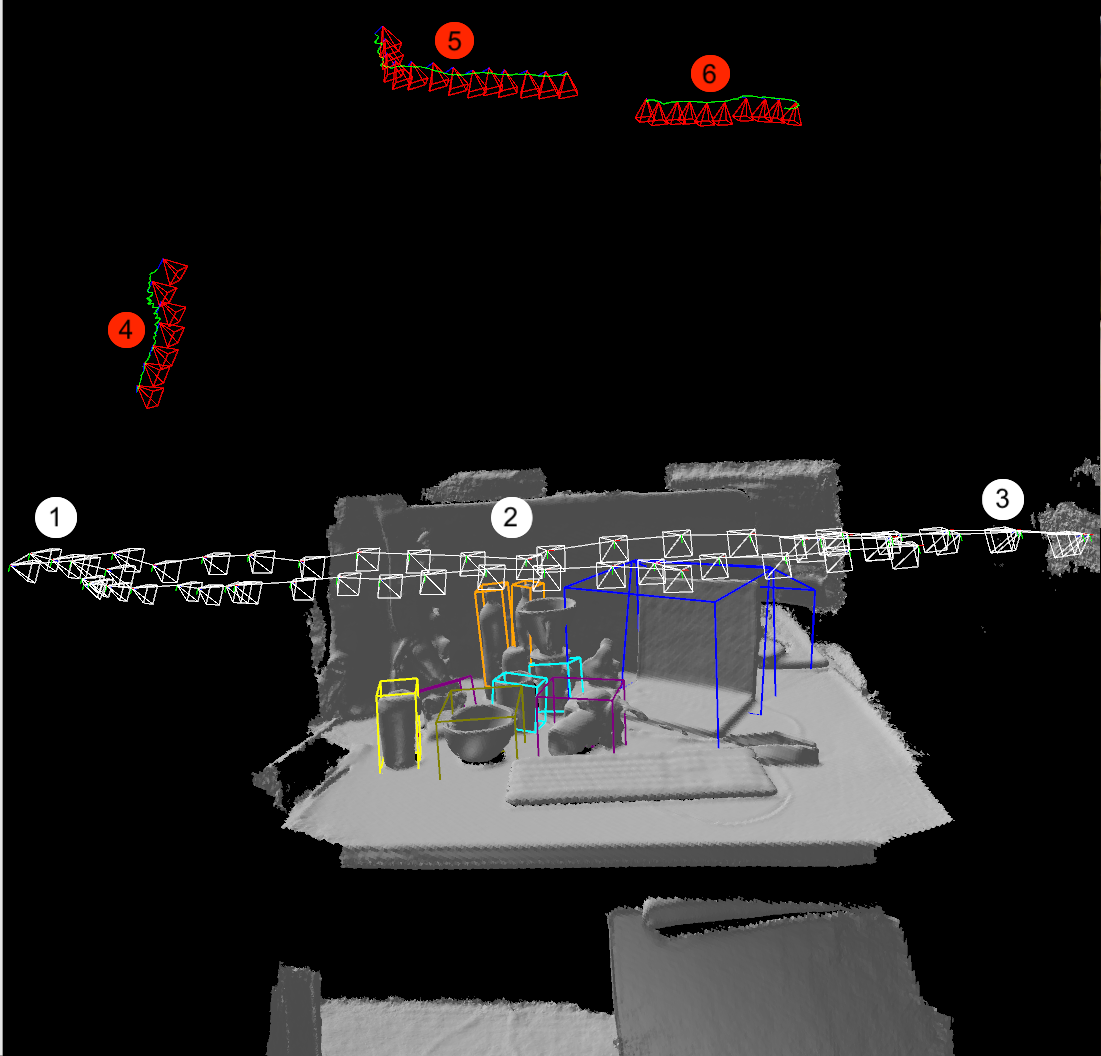} }}
& \multicolumn{3}{l|}{Map construction segment (MCS)} 
\\
\multicolumn{2}{|c}{}
&\includegraphics[width=0.18\textwidth]{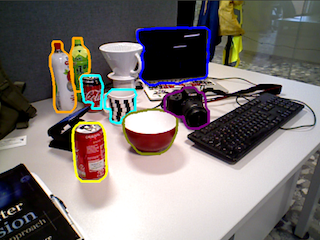}
&\includegraphics[width=0.18\textwidth]{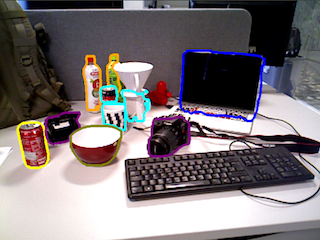}
&\includegraphics[width=0.18\textwidth]{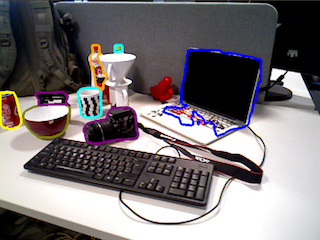}
\\
\multicolumn{2}{|c}{}
&Example 1
&Example 2
&Example 3
\\
\multicolumn{2}{|c}{}
& \multicolumn{3}{l|}{Relocalisation segment (RS)} 
\\
\multicolumn{2}{|c}{}
&\includegraphics[width=0.18\textwidth]{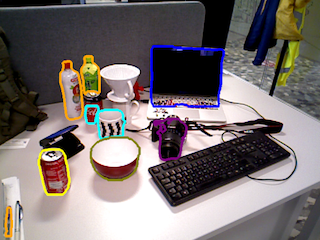}
&\includegraphics[width=0.18\textwidth]{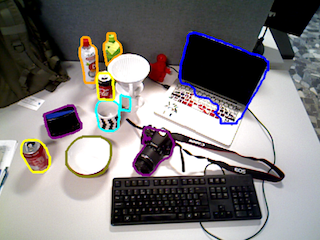}
&\includegraphics[width=0.18\textwidth]{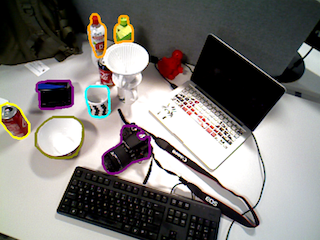}
\\
\multicolumn{2}{|c}{}
&Example 4 ($30^{\circ}(v)$)
&Example 5 ($60^{\circ}(v)$)
&Example 6 ($60^{\circ}(v)$)
\\
\hline
\end{tabular}}
\caption{
Two examples of successful relocalisation for scene 05 and 07 in the OR10 dataset. Left images show the dense partial map, the trajectory for the MCS (white frustums), relocalisation estimates (red frustums) and ground truth trajectories for the RS (green curves). Images to the right show example frames from the MCS and RS, with approximate view change angles shown for the latter. Detected known object categories are highlighted in each and the location of each frame w.r.t the map are shown numbered in the images on the left.
}
\label{fig:result-examples}
\vspace*{-3ex}
\end{figure*}

\subsection{Datasets}
 
We require the test scenes to contain the known objects. Hence, we captured our own RGB-D datasets for ten desktop scenes (OR-10) to test the relocalisation algorithm. In the absence of the ground truth we used our SLAM system to create pseudo ground truth of frame poses. Specifically, for each scene, we first captured a single RGB-D sequence and used this to construct a dense map and estimate the 6D pose of each frame.  We used the latter as the pseudo ground truth and used the partial dense map which was created after an initial segment of the sequence - the {\em map construction segment (MCS)} - as the reference map for relocalisation. The length of this segment varied between the different scenes. We also created three 100-frame segments to test relocalisation - the {\em relocalisation segments (RS)} - each corresponding to different viewing angles around the scene and in some cases sharing frames with the MCS (details below). These were then used as our `lost' frames.

With respect to content complexity (number of objects, degree of clutter, etc), scene 02, 04 and 08 are considered as low, scene 00, 01, 03 and 05 as medium, and scenes 06, 07 and 09 as high. The RS viewpoints were as follows: for scene 00, they were frames from the MCS, which for this scene was the complete original sequence; for scenes 01, 02 and 03, they were taken along a similar trajectory to that for the MCS; and for the remainder they were taken along very different trajectories, with wide horizontal ($h$) or vertical ($v$) view changes. Approximate view changes for the three segments in each case were: scene 04, 05, and 08, $30^{\circ}(h)/120^{\circ}(h)/180^{\circ}(h)$; scene 06, $45^{\circ}(h)/90^{\circ}(h)/135^{\circ}(h)$; scene 07, $30^{\circ}(v)/60^{\circ}(v)/60^{\circ}(v)$; and scene 09, $30^{\circ}(v)/90^{\circ}(h)/180^{\circ}(h)$. Taking both into account, we classified the scenes into four levels of difficulty: easy (scenes 00 and 02); moderate (scenes 01, 03, 04 and 08); difficult (scenes 05 and 07) and very difficult (scenes 06 and 09). Example frames from MCS and RS are shown in Fig. \ref{fig:result-examples}.

\begin{table}[t]
\caption{Relocalisation success rates}
\label{tab:compare}
\vspace*{-3ex}
\begin{center}
\begin{tabular}{|c||c|c|c|c|}
\hline
& Ours & Ours w/o ICP & BoW & FERNS \\
\hline
\hline
scene00 & $\bm{97.00}$\% & 58.00\% & 44.33\% & 74.00\% \\
\hline
\hline
scene01 & $\bm{100}$\%   & 89.67\% & 75.33\% & 83.67\% \\
\hline
scene02 & $\bm{100}$\%   & 71.00\% & 58.67\% & 75.00\% \\
\hline
scene03 & 94.67\%        & 68.67\% & 51.67\% & $\bm{100}$\%  \\
\hline
\hline
scene04 & $\bm{97.67}$\% & 47.33\% & 0.00\%  & 0.00\%  \\
\hline
scene05 & $\bm{89.33}$\% & 70.67\% & 13.67\% & 33.33\% \\
\hline
scene06 & $\bm{72.67}$\% & 42.00\% & 0.00\%  & 13.67\% \\
\hline
scene07 & $\bm{99.33}$\% & 44.00\% & 16.67\% & 49.67\% \\
\hline
scene08 & 50.00\% & $\bm{54.00}$\% & 0.67\%  & 0.00\%  \\
\hline
scene09 & 19.67\%        & 2.00\%  & $\bm{22.00}$\% & 2.33\%  \\
\hline
\end{tabular}
\end{center}
\vspace*{-2ex}
\end{table}

\begin{table}[t]
\caption{Relocalisation performance with/without ICP}
\label{tab:icp}
\vspace*{-3ex}
\begin{center}
\begin{tabular}{|c||c|c|c|}
\hline
& 5cm/$5^{\circ}$ & 10cm/$10^{\circ}$ & 15cm/$15^{\circ}$ \\
\hline
\hline
Ours         & 82.03\% & 87.20\% & 87.83\% \\
\hline
Ours w/o ICP & 54.73\% & 77.73\% & 84.20\%  \\
\hline
\end{tabular}
\end{center}
\vspace*{-4ex}
\end{table}

\subsection{Comparison}
We compared the performance of our method with two appearance based approaches. These were the Bag-of-Words (BoW) method in \cite{Mur2014BoW} and the Randomised Ferns (FERNS) method in \cite{Glocker2013FERNS}. We used the ORB-SLAM2 \cite{Mur2017ORBSLAM2} and InfiniTAMv3  \cite{Prisacariu2017infinitamv3} implementations, respectively. Performance was measured according to the number of frames in the RS that were successfully relocalised. We based the latter on the error metric that the estimated camera pose lie within 5cm translational error and $5^{\circ}$ angular error of the pseudo ground truth, as commonly used by others \cite{Li2015Pairwise, Shotton2013SCoRe}. 
Additionally, we also present the performance of our method without performing the final depth-centroid ICP optimisation step.

\subsection{Results}
Figure \ref{fig:result-examples} shows examples of successful relocalisations. White frustums represent key frames, illustrating the camera trajectory of the MCS. Successfully relocalised `lost' frames are shown as red frustums and the green curves correspond to pseudo ground truth trajectories of the RS. The blue lines connecting the red frustums and green curves illustrate the differences between the estimated and ground truth poses. Selected frames from the MCS and RS are also shown, with numbers indicating their position along each trajectory, as shown in the left-hand map view. We only visualise the result of the proposed method because most of the failed cases of BoW or FERNS produce no result poses due to the significant changes in viewing angles. The experiments are done on a laptop with an AMD Ryzen 7 5800h CPU and a GeForce RTX 3070 Laptop GPU. On average, it takes 0.4267s for object detection, 0.0002s for object-level relocalisation, and 0.0075s for depth-centroid ICP. We recommend readers to watch the attached video to fully appreciate the relocalisation performance.

Table \ref{tab:compare} shows quantitative results for the different methods in terms of relocalisation success rates. Our method has very high success rates, outperforming the other methods for the vast majority of scenes, notably those in which the `lost' frames are taken from very disparate views compared to those used to construct the map. In these cases, the appearance features used by the other two methods are either not visible or have significantly changed, resulting in relocalisation failure. In contrast, object presence is maintained in these disparate views, allowing our method to succeed, illustrating the benefit of the object based approach. Note that although performance drops a little for scene 06, it still surpasses that of the other methods. 

However, for scenes 08 and 09, performance drops considerably. In the case of 08, this was primarily due to significant errors in the reference map, resulting from erroneous RGB-D data caused by reflections from the laptop screen and monitor along the RS trajectories, causing the depth-centroid ICP to converge on incorrect frame poses. For 09, the main reason for failure was inconsistent object estimates provided by the NOCS network. The scene is very cluttered, with many of the known objects partially occluded in most views, leading to highly inconsistent detections in terms of predicted category labels and estimated 6D poses. Similar problems were also noted for 06, but the scene is less cluttered and so the impact on performance was less severe.

In Table \ref{tab:icp}, we compare the average successful rate over all 10 scenes under looser criteria. The result demonstrates that the probabilistic AO can only relocalise the lost frames to the poses within around 10cm and $10^{\circ}$ away from the ground truth, thus making it necessary to further optimise the poses using proposed depth-centroid ICP.

\section{Conclusion and Future Work}
\label{sec:conclusion}
We have presented an object based approach to relocalisation capable of relocalising single frames at viewpoints with significant disparity from those views used to construct the map and for scenes with significant clutter. The compactness of the object  representation also means that it can be scaled to very large maps. For future work, we intend to investigate techniques for making use of previously unseen objects and how better to deal with scenes containing greater levels of clutter, as well as more in-depth comparison with recent deep learning approaches to 3-D point cloud matching.

\end{document}